\documentclass{article}
\usepackage{ijcai26}

\usepackage{multirow}
\usepackage{times}
\usepackage{soul}
\usepackage{url}
\usepackage[hidelinks]{hyperref}
\usepackage[utf8]{inputenc}
\usepackage[small]{caption}
\usepackage{graphicx}
\usepackage{amsmath}
\usepackage{amsthm}
\usepackage{booktabs}
\usepackage{algorithm}
\usepackage{algorithmic}
\usepackage[switch]{lineno}

\usepackage{tabularx}
\usepackage{array}
\newcolumntype{Y}{>{\raggedright\arraybackslash}X}

\usepackage{amssymb}

\usepackage{enumitem}
\usepackage{placeins}

\title{Personalized Federated Sparse Adaptation of Time-Series Foundation Models\footnote{Accepted at International Workshop on Federated Learning in the Age of Foundation Models
In Conjunction with IJCAI 2026 (FL@FM-IJCAI'26)}}

\author{
Priyanka Nihalchandani$^{1}$\and
Naman Srivastava$^{1}$\and
Varun Ojha$^{2}$\and
Pandarasamy Arjunan$^{1}$\\
\affiliations
$^{1}$Robert Bosch Centre for Cyber-Physical Systems, Indian Institute of Science, Bengaluru, India\\
$^{2}$School of Computing, Newcastle University, Newcastle upon Tyne, United Kingdom\\
\emails
priyankan@iisc.ac.in, snaman@iisc.ac.in, varun.ojha@newcastle.ac.uk, samy@iisc.ac.in
}

\begin{document}

\maketitle

\begin{abstract}
Federated adaptation of time-series foundation models (TSFMs) is attractive for
building energy forecasting because meter data are private, distributed, and
highly non-IID. However, a single parameter-sharing strategy is unlikely to
serve all pretrained TSFMs or building clients: fully shared adapters can
suppress building-specific temporal behavior, while fully local adaptation
discards cross-building transfer. We propose a personalized federated sparse adaptation framework with a
heterogeneous temporal mixture-of-experts (MoE) adapter placed after the
pretrained TSFM representation. A sequence-level router assigns each input window to a sparse subset
of experts capturing complementary temporal inductive biases. We compare global FL, local training,
and personalized FL variants with globally shared or client-private expert
banks. Across 50 buildings and three TSFM backbones, personalized adaptation
consistently improves over Global FL-MoE and is generally competitive with
Local MoE. The strongest adaptation strategy varies across
backbones and evaluation metrics, highlighting the need for
backbone-aware personalization. Across the three TSFMs, the best-performing personalized variants reduce NRMSE by
8.2\%, 7.1\%, and 12.5\% relative to Global FL-MoE for MOMENT,
Chronos-2, and Moirai, while simultaneously reducing communication through selective parameter sharing. Routing analysis reveals backbone-specific expert utilization patterns, including client-level specialization in MOMENT, expert concentration in Chronos-2, and more uniform expert usage in Moirai.
\end{abstract}

\section{Introduction}

Time-series foundation models (TSFMs) have shown strong transfer ability for
forecasting by learning reusable temporal representations from large-scale
pretraining corpora~\cite{goswami2024moment,ansari2025chronos2univariateuniversalforecasting,Li2025TSFMBench,woo2024unified}.
Building energy forecasting is a natural target for TSFMs because short-term
load prediction supports demand response, HVAC operation, fault detection, and
grid-aware control~\cite{saravanan2024analyzing,Prabhu2022EPTK,Prabhu2023XAIEnergy}.
However, building-level meter data are distributed across facilities and are
often difficult to centralize because of privacy, ownership, and operational
constraints. Federated learning (FL) offers a way to adapt forecasting models
across buildings without sharing raw time-series data
~\cite{pmlr-v54-mcmahan17a,NEURIPS2024_abc19438,Chen2025FFTS}.

Federated TSFM adaptation is challenging because building clients are highly
non-IID: they differ in load scale, occupancy rhythms, operating schedules,
seasonal sensitivity, and transient usage patterns. Fully shared adaptation can
average away these local temporal signatures, whereas fully local adaptation
forgoes cross-building transfer. This motivates personalized parameter-efficient
adaptation, where selected components are shared through FL while others remain
client-specific
~\cite{MLSYS2020_1f5fe839,hu2022lora,zhang2023adaptive}.

We study personalization through sparse expert adaptation. Rather than
personalizing only prediction heads or shallow layers
~\cite{arivazhagan2019federatedlearningpersonalizationlayers,oh2022fedbabu,NEURIPS2020_f4f1f13c},
we route each input window through a heterogeneous temporal expert bank that
captures complementary inductive biases, including periodicity, long-range
interaction, local temporal variation, trend--residual structure, and
multi-resolution behavior
~\cite{NIPS2017_3f5ee243,wu2021autoformer,pmlr-v162-zhou22g}. Beyond improving
forecasting performance, expert-selection patterns provide insight into how
different pretrained TSFM backbones organize building-specific temporal
behavior.

We propose a personalized federated sparse adaptation framework and evaluate it
across MOMENT-1-large, Chronos-2, and Moirai-1.1-R-small on 50
non-residential buildings. We compare Global FL-MoE, Local MoE, PFL without
MoE, shared-expert PFL-MoE, and private-expert PFL-MoE. Our results show that
personalized adaptation consistently improves over Global FL-MoE and is
generally competitive with or better than Local MoE. More importantly, the
optimal adaptation strategy varies across TSFM backbones and evaluation
metrics, indicating that federated personalization should be designed jointly
with the underlying pretrained model rather than treated as a backbone-agnostic
procedure. Routing analysis further reveals distinct backbone-specific expert
utilization regimes, including client-level specialization in MOMENT, expert
concentration in Chronos-2, and near-uniform routing in Moirai.

Our contributions are:
\begin{itemize}[leftmargin=*,nosep]

\item We propose a personalized federated sparse-adaptation framework for
adapting pretrained TSFMs under non-IID building clients without sharing raw
meter data.

\item We design a post-representation temporal MoE adapter with sequence-level
top-$k$ routing over heterogeneous experts that capture periodic, long-range,
local, trend--residual, and multi-resolution temporal patterns.

\item We systematically compare global, local, non-MoE personalized,
shared-expert, and private-expert adaptation strategies across 50 buildings and
three TSFM backbones.

\item We show that the effectiveness of federated personalization is
backbone-dependent and characterize the resulting communication--accuracy
trade-offs and expert-utilization patterns through extensive empirical and
statistical evaluation.
\end{itemize}

\section{Related Work}

\paragraph{Time-series foundation models.}
Time-series foundation models (TSFMs) learn transferable temporal representations through large-scale pretraining and have demonstrated strong zero-shot and fine-tuning performance across diverse forecasting tasks. Representative models include MOMENT, Chronos, and Moirai, which adopt different pretraining objectives and architectural designs for general-purpose time-series forecasting~\cite{goswami2024moment,ansari2025chronos2univariateuniversalforecasting,woo2024unified}. Recent benchmarks have systematically evaluated TSFMs across zero-shot, few-shot, and full-shot forecasting settings~\cite{Li2025TSFMBench}. Their potential for building-energy applications has also attracted growing interest, including predictive building analytics and short-term load forecasting~\cite{Mulayim2024TSFMBuildingAnalytics,saravanan2024analyzing}.

\paragraph{Federated and personalized adaptation.}
Federated learning enables collaborative model adaptation without sharing raw data, but standard FedAvg can degrade under heterogeneous client distributions~\cite{pmlr-v54-mcmahan17a,MLSYS2020_1f5fe839}. Personalized FL methods address this challenge by separating globally shared representations from client-specific parameters through private heads, personalization layers, final personalization stages, or regularized local objectives~\cite{arivazhagan2019federatedlearningpersonalizationlayers,oh2022fedbabu,NEURIPS2020_f4f1f13c}. Recent work has begun extending these ideas to foundation models and time-series forecasting, including LM-empowered federated TSFMs and heterogeneous TSFM training~\cite{NEURIPS2024_abc19438,Chen2025FFTS}. More recent approaches such as pFedDKS further explore selective knowledge sharing mechanisms that balance global transfer and client-specific adaptation while preserving privacy~\cite{10.1145/3774904.3792203}. However, it remains unclear how sparse expert adapters for pretrained TSFMs should be shared or personalized under federated non-IID building clients.

\paragraph{Sparse experts and temporal inductive biases.}
Sparse mixture-of-experts (MoE) architectures route each input to a subset of expert modules, enabling conditional computation and expert specialization~\cite{lepikhin2021gshard,JMLR:v23:21-0998}. MoE-based forecasting has been explored for energy applications under federated learning, where expert gating improves adaptation to heterogeneous load, PV, and prosumption time series~\cite{sievers2024advancing}. Recent TSFM research further demonstrates that sparse experts can specialize to diverse temporal patterns~\cite{liu2025moiraimoe}. In contrast to generic expert banks, our approach combines experts encoding temporal inductive biases that are particularly relevant to building-energy forecasting, including periodicity, long-range dependency, local temporal variation, trend--residual decomposition, and multi-resolution structure. Beyond forecasting performance, we analyze routing behavior to understand how different TSFM backbones and building clients utilize these temporal mechanisms under federated personalization.

\section{Methodology}

\subsection{Problem Formulation}
\label{sec:problem}

Let $\mathcal{C}=\{1,\ldots,N\}$ denote a set of building clients. Each client
$i\in\mathcal{C}$ owns a private hourly electricity-consumption time series
$\mathbf{x}^{(i)}\in\mathbb{R}^{T_i}$. Client data are heterogeneous and
non-IID, with building-specific distributions $\mathcal{P}_i$ that may differ
across clients. Raw time-series data are never shared with the server.

Given a context window of length $L=168$ hours, each client learns a forecasting
function
\[
f^{(i)}:\mathbb{R}^{L}\rightarrow\mathbb{R}^{H},
\]
where $H=24$ is the prediction horizon. We adapt a pretrained time-series
foundation model (TSFM) under federated learning and study three backbones:
MOMENT-1-large, Chronos-2, and Moirai-1.1-R-small.

\begin{figure*}[!t]
    \centering
    \includegraphics[width=\linewidth]{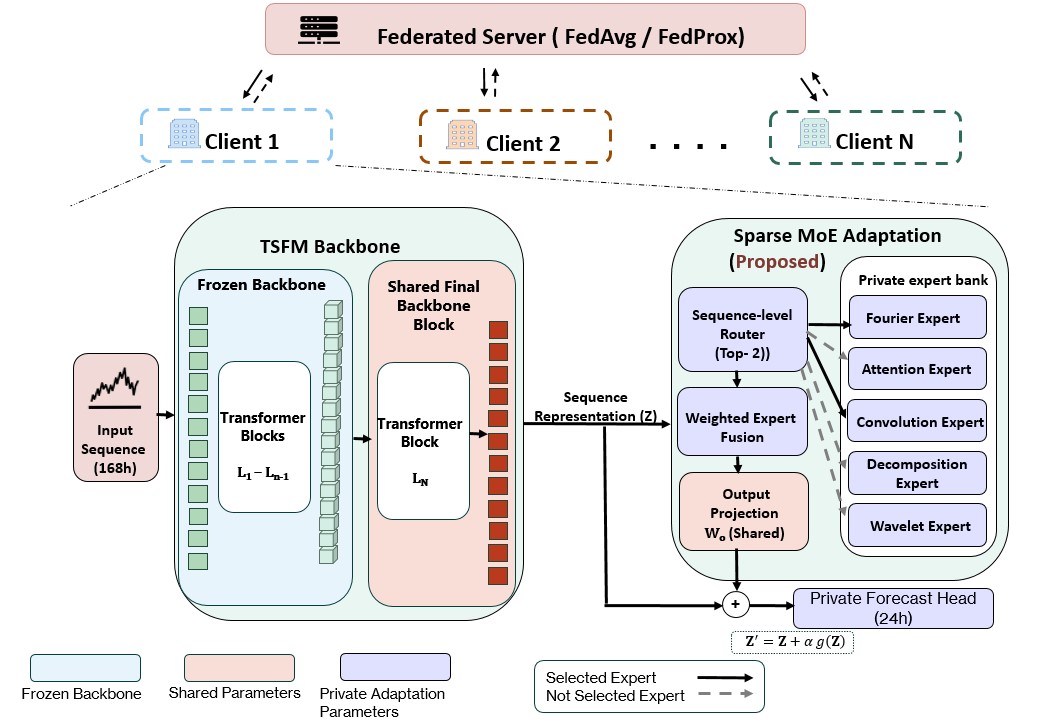}
    \caption{
Overview of the proposed personalized federated sparse-adaptation framework
(private-expert PFL-MoE variant). A pretrained TSFM encodes each
input into hidden states $\mathbf{Z}$. A client-private sequence-level router
selects top-$k$ experts from a client-private temporal expert bank; selected
outputs are fused, projected by the shared output projection $\mathbf{W}_o$,
and added residually before the private forecasting head. Colors denote frozen,
shared, and private components.
}
    \label{fig:architecture}
\end{figure*}

\subsection{Post-Representation Sparse MoE Adapter}
\label{sec:adapter}

Each pretrained TSFM is used as a representation-producing backbone. Given a
batch of input context windows, the backbone maps the inputs to a sequence
of hidden representations:
\begin{equation}
    \mathbf{Z}
    =
    \mathcal{F}_{\boldsymbol{\phi}}(\mathbf{x})
    \in \mathbb{R}^{B\times P\times D},
    \label{eq:hidden}
\end{equation}
where $B$ is the batch size, $P$ is the number of temporal patches or
hidden tokens, and $D$ is the hidden dimension. Most pretrained backbone parameters remain frozen to reduce
communication and local optimization cost. We introduce a lightweight sparse mixture-of-experts
(MoE) adapter after the final backbone representation and before the
backbone-specific forecasting or distribution head.

The adapter produces an adapted representation through a residual update:
\begin{equation}
\mathbf{Z}' = \mathbf{Z} + \alpha g(\mathbf{Z}),
\end{equation}
where $g(\mathbf{Z})$ is the sparse MoE transformation and
$\alpha=\sigma(\ell_\alpha)$ is a learnable scalar residual gate. We optimize
the unconstrained logit $\ell_\alpha$ and use the sigmoid only to keep the
adapter scale bounded. The gate is initialized conservatively so that the adapter initially acts as a residual correction and
does not abruptly change the pretrained backbone representation.

This post-representation placement, illustrated in Figure~\ref{fig:architecture}, provides a common adaptation interface across heterogeneous TSFM backbones. Rather than modifying backbone-specific internal layers, the MoE branch operates on the final hidden representation and produces a residual update that specializes the forecasting representation. In our federated setting, most pretrained backbone parameters remain frozen. The trainable components consist of the final backbone-side adaptation block and the lightweight post-representation adaptation modules. The final backbone-side block is always shared through FL, while the remaining trainable components are assigned to shared or client-private parameter sets depending on the personalization strategy being evaluated. This design enables backbone-agnostic adaptation while reducing communication and local optimization costs by limiting the number of trainable parameters.

\subsection{Sequence-Level Sparse Routing and Expert Fusion}
\label{sec:routing_fusion}

We use sequence-level sparse routing to make expert selection stable at the
forecast-window level. The router makes one routing decision per input sequence. The
same selected expert subset is then used for all hidden tokens in that sequence.
This design is important for building energy forecasting because the relevant
temporal behavior of a 168-hour context window, such as daily periodicity,
local transients, or trend shifts, usually characterizes the whole input window
rather than an isolated patch. Sequence-level routing therefore reduces
patch-wise routing noise and makes the selected experts easier to interpret as
client- and sequence-level temporal preferences.

Given hidden representations
$\mathbf{Z}\in\mathbb{R}^{B\times P\times D}$, the router computes one
sequence-level summary for each input window by pooling over the observed
context-side hidden tokens:
\begin{equation}
    \bar{\mathbf{z}}_b
=
\frac{1}{|\mathcal{I}_b|}
\sum_{p\in\mathcal{I}_b}
\mathrm{LN}_{r}(\mathbf{Z}_{b,p,:}),
\qquad
\bar{\mathbf{Z}}\in\mathbb{R}^{B\times D}.
    \label{eq:seq_summary}
\end{equation}
Here, $b\in{1,\ldots,B}$ indexes one input sequence within the batch, and
$\mathcal{I}_b$ denotes the set of observed context-token indices for that
sequence. For backbones whose hidden sequence contains only context-side tokens,
$\mathcal{I}_b={1,\ldots,P}$, so Eq.~\eqref{eq:seq_summary} reduces to
mean-pooling over all hidden tokens. For backbones whose native forecasting
interface includes prediction-side hidden tokens, routing is based only on the
observed context tokens so that expert selection does not depend on future-side
positions.

Router logits and probabilities are computed as
\begin{equation}
    \boldsymbol{\ell}
    =
    \bar{\mathbf{Z}}\mathbf{W}_{r}^{\top},
    \qquad
    \boldsymbol{\pi}
    =
    \mathrm{softmax}(\boldsymbol{\ell}),
    \label{eq:router}
\end{equation}
where $\boldsymbol{\ell},\boldsymbol{\pi}\in\mathbb{R}^{B\times M}$,
$\mathbf{W}_{r}\in\mathbb{R}^{M\times D}$ is the router matrix, and $M$ is the
number of experts.

For each sequence $b$ in the batch, the router selects the top-$k$ experts:
\begin{equation}
    \mathcal{T}_{b}
    =
    \mathrm{top\text{-}}k(\boldsymbol{\ell}_{b,:}),
    \qquad
    |\mathcal{T}_{b}|=k.
    \label{eq:topk}
\end{equation}
Top-$k$ selection is applied independently to each input sequence. Since softmax preserves ordering, selecting top-$k$ logits
is equivalent to selecting top-$k$ routing probabilities. In the main
sparse-execution setting, only the selected experts are executed. We use $k=2$
as the main sparse-routing setting and vary $k$ in routing-sparsity ablations. Top-2 routing provides sparse expert execution while allowing multiple complementary temporal inductive biases to contribute to a single forecast window. In contrast, top-1 routing enforces harder specialization, whereas top-5 routing effectively removes sparsity by activating all experts.

The main expert bank contains five heterogeneous temporal experts: Fourier,
attention, convolution, decomposition, and wavelet experts. These
experts are not intended to be novel temporal modules individually; rather,
their role is to provide a compact set of complementary inductive biases for
personalized TSFM adaptation. The Fourier expert targets periodic structure,
the attention expert captures long-range patch interactions, the
convolution expert models local and multi-scale temporal variation, the
decomposition expert separates trend and residual components, and the wavelet
expert captures localized time-frequency behavior. Each expert
$E_m$ maps $\mathbb{R}^{B \times P \times D}$ to
$\mathbb{R}^{B \times P \times D}$. We also evaluate reduced expert-bank
variants as ablations.

For selected experts $m\in\mathcal{T}_{b}$, routing weights are normalized over
the selected set:
\begin{equation}
    \tilde{\pi}_{b,m}
    =
    \frac{\exp(\ell_{b,m})}
    {\sum_{m'\in\mathcal{T}_{b}}\exp(\ell_{b,m'})},
    \qquad
    m\in\mathcal{T}_{b}.
    \label{eq:renorm}
\end{equation}

The sparse MoE transformation for sequence $b$ is
\begin{equation}
    g(\mathbf{Z})_{b,:,:}
=
\mathbf{W}_{o}
\left(
    \sum_{m\in\mathcal{T}_{b}}
    \tilde{\pi}_{b,m}
    E_{m}(\mathbf{Z}_{b,:,:})
\right).
    \label{eq:moe_transform}
\end{equation}
where $\mathbf{W}_{o}$ is the adapter output projection. We keep $W_o$ shared to provide a common latent adaptation interface across clients, while allowing expert transformations and routing decisions to be personalized. The adapted
representation $\mathbf{Z}'$ is then passed to the backbone-specific forecasting
or distribution head; in personalized variants, this head is client-private.
\paragraph{Expert-selection analysis.}
On held-out test windows, we compute, for each client, the fraction of windows
in which each expert is selected by the top-$k$ router. These selection
patterns indicate whether a backbone uses client-specific experts, concentrates
on a small set of experts, or spreads selection across the expert bank.

\subsection{Federated Objective and Training}
\label{sec:objective_training}
The personalized federated objective is:
\begin{equation}
    \min_{\boldsymbol{\theta}_{s},
    \{\boldsymbol{\theta}_{p}^{(i)}\}_{i=1}^{N}}
    \sum_{i=1}^{N}
    \frac{n_i}{\sum_{j=1}^{N}n_j}
    \mathcal{L}_{i}
    \left(
    \boldsymbol{\theta}_{s},
    \boldsymbol{\theta}_{p}^{(i)}
    \right).
    \label{eq:objective}
\end{equation}
where $\boldsymbol{\theta}_{s}$ denotes trainable shared parameters and
$\boldsymbol{\theta}_{p}^{(i)}$ denotes client-private parameters for client
$i$. Here, $n_i$ denotes the number of training windows available at client $i$. The exact shared/private split depends on the adaptation setting defined in
Section~\ref{sec:experiments}.

For MoE-based variants, each selected client minimizes
\begin{equation}
    \mathcal{L}_{\mathrm{total}}
    =
    \mathcal{L}_{\mathrm{forecast}}
    +
    \lambda_{\mathrm{lb}}\mathcal{L}_{\mathrm{lb}}
    +
    \lambda_{z}\mathcal{L}_{z},
    \label{eq:loss}
\end{equation}
where $\mathcal{L}_{\mathrm{forecast}}$ is the backbone-specific forecasting
loss, and $\mathcal{L}_{\mathrm{lb}}$ and $\mathcal{L}_{z}$ are standard sparse
MoE load-balancing and router regularization losses
~\cite{JMLR:v23:21-0998,Zoph2022STMoEDS,lepikhin2021gshard}.

\begin{equation}
    \mathcal{L}_{\mathrm{lb}} = M \sum_{m=1}^{M} I_m P_m,
\end{equation}

Here, $I_m$ denotes the fraction of sequences in the batch for which expert
$m$ is selected by the top-$k$ router, and $P_m$ denotes the average routing
probability assigned to expert $m$ over the batch.

\begin{equation}
    \mathcal{L}_{z} = \frac{1}{B}\sum_{b=1}^{B}
\left(\log \sum_{m=1}^{M}\exp(\ell_{b,m})\right)^2.
\end{equation}

The forecasting loss follows each backbone's native training interface: MSE for
MOMENT, quantile pinball loss for Chronos-2, and masked negative log-likelihood
for Moirai.

Training proceeds for $T$ communication rounds with partial client
participation. At each round, selected clients update both shared and
client-private parameters locally. The server aggregates only the uploaded
shared parameters using sample-weighted FedAvg~\cite{pmlr-v54-mcmahan17a}. We
also evaluate FedProx~\cite{MLSYS2020_1f5fe839}, where the proximal penalty is
applied only to shared parameters. After the final communication round, each
client performs final personalization by freezing $\boldsymbol{\theta}_{s}^{T}$
and updating only its private parameters $\boldsymbol{\theta}_{p}^{(i)}$.

\begin{algorithm}[!t]
\caption{Personalized Federated Sparse-MoE Adaptation}
\label{alg:pfl_moe}
\begin{algorithmic}[1]
\STATE \textbf{Input:} clients $\mathcal{C}$, rounds $T$, local epochs
$E_{\mathrm{local}}$, final personalization epochs $E_{\mathrm{final}}$,
participation rate $\rho$, partition
$q\in\{\mathrm{shared\text{-}experts},\mathrm{private\text{-}experts}\}$,
routing sparsity $k$, optimizer $a$, selection rule $s$
\STATE Initialize frozen backbone $\boldsymbol{\theta}_f$, shared parameters
$\boldsymbol{\theta}_s^0$, and private parameters
$\{\boldsymbol{\theta}_p^{(i),0}\}_{i=1}^{N}$ according to $q$
\FOR{$t=1,\ldots,T$}
    \STATE Select $\mathcal{S}^t\subset\mathcal{C}$ using $s$, with
    $|\mathcal{S}^t|=\lceil\rho N\rceil$
    \FOR{each client $i\in\mathcal{S}^t$ in parallel}
        \STATE Load $\boldsymbol{\theta}_s^{t-1}$ and
        $\boldsymbol{\theta}_p^{(i),t-1}$
        \STATE Update $(\boldsymbol{\theta}_s,\boldsymbol{\theta}_p^{(i)})$
        for $E_{\mathrm{local}}$ epochs using
        $\mathcal{L}_{\mathrm{total}}$ with
        $\mathbf{Z}'=\mathbf{Z}+\alpha g(\mathbf{Z})$
        \IF{$a=\mathrm{FedProx}$}
            \STATE Add $\frac{\mu}{2}
            \|\boldsymbol{\theta}_s-\boldsymbol{\theta}_s^{t-1}\|_2^2$
            only on shared parameters
        \ENDIF
        \STATE Upload $\boldsymbol{\theta}_s^{(i),t}$ and keep
        $\boldsymbol{\theta}_p^{(i),t}$ local
    \ENDFOR
    \STATE Aggregate
    $\boldsymbol{\theta}_s^t=
    \sum_{i\in\mathcal{S}^t}
    \frac{n_i}{\sum_{j\in\mathcal{S}^t}n_j}
    \boldsymbol{\theta}_s^{(i),t}$
\ENDFOR
\FOR{each client $i\in\mathcal{C}$ in parallel}
    \STATE Freeze $\boldsymbol{\theta}_s^T$ and update only
    $\boldsymbol{\theta}_p^{(i),\star}$ for $E_{\mathrm{final}}$ epochs
\ENDFOR
\STATE \textbf{return} $\boldsymbol{\theta}_s^T$ and
$\{\boldsymbol{\theta}_p^{(i),\star}\}_{i=1}^{N}$
\end{algorithmic}
\end{algorithm}
Algorithm~\ref{alg:pfl_moe} summarizes the personalized MoE variants. Global
FL-MoE, Local MoE, and PFL without MoE use the same protocol with the parameter
partition or MoE branch changed as described in Section~\ref{sec:experiments}.
In the shared-expert variant, the expert bank and $\mathbf{W}_o$ are shared,
while the router, residual gate, and forecasting head remain private. In the private-expert variant, the expert bank is also client-private, while
$\mathbf{W}_o$ remains shared. In Algorithm~\ref{alg:pfl_moe}, $\mu$ denotes
the FedProx proximal coefficient.


\section{Experiments and Results}
\label{sec:experiments}

\paragraph{Dataset and task.}
We evaluate on a subset of the ASHRAE Great Energy Predictor III dataset ~\cite{miller2020building} 
containing hourly electricity-consumption data from 50 non-residential buildings. This setting provides a controlled but heterogeneous cross-building FL benchmark
in which each client has a distinct load distribution.
Each building is treated as one federated client, yielding a non-IID
cross-building forecasting setting. For each client, the time series is split
chronologically into 50\% training, 25\% validation, and 25\% testing data. A
standard scaler is fitted only on the training split of each client and applied
to the corresponding validation and test splits. The forecasting task uses a 168\,hour context window to predict the next 24\,hours. During evaluation, the test split is partitioned into consecutive non-overlapping 24\,hour forecast windows, and metrics are computed after transforming predictions back to the original energy scale.

\paragraph{Backbones and adapter configuration.}
We evaluate three pretrained TSFM backbones: MOMENT-1-large, Chronos-2, and
Moirai-1.1-R-small. Across all backbones, the adapter is applied after the final
hidden representation and before the backbone-specific forecasting or
distribution head. Unless otherwise stated, PFL-MoE uses the full five-expert
bank \{Fourier, Attention, Convolution, Decomposition, Wavelet\},
sequence-level top-$k=2$ routing, and sparse expert execution. The parameter-sharing choices for each adaptation setting are defined under
Baselines and personalized variants. Appendix~\ref{app:hyperparameters} summarizes the shared training and
federated-learning hyperparameters used across experiments, including
communication rounds, participation rates, routing sparsity,
regularization coefficients, and optimization settings.

\paragraph{Federated training protocol.}
We train for 15 communication rounds with 20\% client participation per round.
Each selected client performs 5 local epochs before server aggregation. Unless
otherwise stated, shared parameters are aggregated using sample-weighted
FedAvg, and clients are selected using a coverage-based strategy that
prioritizes under-selected clients under partial participation. We compare
random client selection and FedProx as federated-training ablations. We maintain a participation count for each client across communication rounds. At each round, clients with lower historical participation counts are assigned higher selection priority to encourage balanced coverage under partial participation. When multiple clients have the same participation count, ties are broken uniformly at random. This strategy ensures that all clients are selected regularly while preserving stochasticity in client participation.

\paragraph{Baselines and personalized variants.}
We compare the following adaptation settings:
\begin{itemize}[leftmargin=*,nosep]
    \item \textbf{Zero-shot TSFM:} the pretrained backbone is evaluated without
    any fine-tuning.
    
    \item \textbf{Global FL-MoE:} all trainable adapter, router, residual gate,
    and forecasting-head parameters are globally aggregated, with no
    client-private state.
    
    \item \textbf{Local MoE:} each building trains its own MoE adapter locally
    without communication.
    
    \item \textbf{PFL without MoE:} we remove the sparse expert branch and retain
the same personalized FL protocol. The trainable backbone-side adaptation
parameters are globally shared, while the forecasting head remains
client-private; no router, expert bank, residual MoE gate, or MoE output
projection is used. This baseline isolates the effect of personalized federated adaptation without conditional expert routing.
    
    \item \textbf{Shared-expert PFL-MoE:} the expert bank and adapter output
    projection $\mathbf{W}_o$ are globally shared, while the router, residual
    gate, and forecasting head remain client-private. The shared-expert variant uses the same architecture, but the expert bank is moved from the private parameter set to the shared parameter set.
    
    \item \textbf{Private-expert PFL-MoE:} the expert bank is also kept
    client-private, while $\mathbf{W}_o$ remains globally shared.
\end{itemize}

Table~\ref{tab:parameter_partitioning} summarizes the
shared and client-private trainable components across all adaptation strategies.

\paragraph{Ablations.}
We evaluate ablations over expert-bank composition, routing sparsity,
federated optimization, client participation, and LoRA adaptation. Four-expert
variants remove one expert from the full expert bank; F, A, C, D, and W denote
Fourier, Attention, Convolution, Decomposition, and Wavelet experts.
Routing ablations compare top-$k \in \{1,2,5\}$ selection, where top-$2$ is the
main sparse setting and top-$5$ selects all experts. LoRA rows denote rank-$8$
low-rank adapter baselines without the MoE branch.

\paragraph{Evaluation metrics.}
We report NRMSE and sMAPE as primary metrics. NRMSE is computed over complete
24-hour forecast windows~\cite{saravanan2024analyzing}.
\begin{equation}
\mathrm{NRMSE}
=
\frac{100}{\bar{y}}
\sqrt{
\frac{1}{24N_{\mathrm{win}}}
\sum_{r=1}^{N_{\mathrm{win}}}
\sum_{h=1}^{24}
\left(y_{r,h}-\hat{y}_{r,h}\right)^2
},
\label{eq:nrmse}
\end{equation}
where $N_{\mathrm{win}}$ is the number of 24-hour forecast windows and $\bar{y}$ is the mean ground-truth electricity consumption over the evaluated windows.

sMAPE is computed over all evaluated forecast time steps as

\begin{equation}
\mathrm{sMAPE}
=
\frac{100}{N}
\sum_{t=1}^{N}
\frac{2\left|y_t-\hat{y}_t\right|}
{\left|y_t\right|+\left|\hat{y}_t\right|+\varepsilon},
\end{equation}

where $N$ denotes the total number of evaluated forecast points and
where $\varepsilon = 10^{-8}$ is a small constant used for numerical stability.

We treat private-expert PFL-MoE with FedAvg, coverage-based client selection,
top-$k=2$, and the full five-expert bank as the default proposed setting. Most ablation rows modify one component relative to the default private-expert
PFL-MoE, while additional shared-expert rows evaluate alternative optimizer or
client-selection settings under the shared-expert partition. In Table~\ref{tab:complete_ablation}, PFL-MoE denotes our default
private-expert personalized sparse-adaptation method. Zero-shot TSFM,
Global FL-MoE, Local MoE, PFL without MoE, and LoRA variants are baselines,
while rows with modified expert banks, routing sparsity, optimizer, client
selection, or shared experts are ablations.

\begin{table}[!h]
\centering
\scriptsize
\setlength{\tabcolsep}{3.2pt}
\renewcommand{\arraystretch}{0.82}
\caption{
Forecasting and ablation results across all TSFM backbones.
NRMSE and sMAPE are reported as median values across 50 building clients
(lower is better). The default proposed method is private-expert PFL-MoE
with FedAvg, coverage-based client selection, top-$k=2$ routing, and the
full five-expert bank. F, A, C, D, and W denote Fourier, Attention,
Convolution, Decomposition, and Wavelet experts. The best result for each backbone and metric is shown in bold, while the second-best result is underlined.
}
\label{tab:complete_ablation}
\begin{tabular}{llcc}
\toprule
\textbf{TSFM} & \textbf{Adaptation / Ablation Setting} &
\textbf{NRMSE} $\downarrow$ & \textbf{sMAPE} $\downarrow$ \\
\midrule

\textbf{MOMENT}
& Private-expert PFL-MoE (top-$1$ routing) & \textbf{12.769} & 9.553 \\
& PFL without MoE & \underline{12.890} & 9.585 \\
& Shared-expert PFL-MoE & 12.893 & 9.565 \\
& Shared-expert PFL-MoE (random CS) & 12.928 & 9.500 \\
& FedProx Shared-expert PFL-MoE & 12.948 & 9.738 \\
& FedProx Private-expert PFL-MoE & 12.975 & 9.630 \\
& Private-expert PFL-MoE (4 experts: F,A,D,W) & 13.015 & \underline{9.408} \\
& Private-expert PFL-MoE (top-$5$ routing) & 13.036 & 9.468 \\
& PFL-LoRA & 13.042 & 9.775 \\
& Local LoRA & 13.048 & 9.439 \\
& Private-expert PFL-MoE (4 experts: F,C,D,W) & 13.072 & 9.497 \\
& Private-expert PFL-MoE (default) & 13.103 & \textbf{9.403} \\
& Local MoE & 13.118 & 9.688 \\
& Private-expert PFL-MoE (4 experts: F,A,C,W) & 13.146 & 9.463 \\
& Private-expert PFL-MoE (4 experts: F,A,C,D) & 13.190 & 9.538 \\
& Private-expert PFL-MoE (4 experts: A,C,D,W) & 13.200 & 9.566 \\
& FedProx Global FL-MoE & 13.747 & 10.355 \\
& Global FL-MoE & 13.912 & 10.349 \\
& Global FL-LoRA & 14.290 & 10.926 \\
& Zero-shot TSFM & 32.275 & 28.112 \\

\midrule

\textbf{Chronos-2}
& Private-expert PFL-MoE (4 experts: F,A,D,W) & \textbf{10.559} & 6.924 \\
& Zero-shot TSFM & \underline{10.573} & 7.144 \\
& Private-expert PFL-MoE (4 experts: F,A,C,D) & 10.598 & \textbf{6.791} \\
& Private-expert PFL-MoE (default) & 10.662 & 6.873 \\
& Private-expert PFL-MoE (4 experts: A,C,D,W) & 10.688 & \underline{6.804} \\
& Private-expert PFL-MoE (4 experts: F,A,C,W) & 10.732 & 6.822 \\
& Private-expert PFL-MoE (top-$5$ routing) & 10.733 & 6.922 \\
& Private-expert PFL-MoE (4 experts: F,C,D,W) & 10.776 & 6.809 \\
& Private-expert PFL-MoE (top-$1$ routing) & 10.803 & 6.855 \\
& Shared-expert PFL-MoE & 10.834 & 6.941 \\
& Shared-expert PFL-MoE (random CS) & 10.898 & 7.056 \\
& FedProx Private-expert PFL-MoE & 10.937 & 6.981 \\
& PFL-LoRA & 10.960 & 7.303 \\
& FedProx Shared-expert PFL-MoE & 10.973 & 7.023 \\
& Local LoRA & 10.987 & 7.102 \\
& PFL without MoE & 10.992 & 6.885 \\
& Local MoE & 11.215 & 7.268 \\
& FedProx Global FL-MoE & 11.259 & 7.838 \\
& Global FL-MoE & 11.360 & 7.764 \\
& Global FL-LoRA & 11.363 & 8.151 \\

\midrule

\textbf{Moirai}
& FedProx Private-expert PFL-MoE & \textbf{12.429} & 8.502 \\
& Private-expert PFL-MoE (4 experts: F,A,C,W) & \underline{12.430} & \underline{8.489} \\
& Private-expert PFL-MoE (default) & 12.610 & 8.634 \\
& Private-expert PFL-MoE (4 experts: F,A,D,W) & 12.623 & 8.758 \\
& Private-expert PFL-MoE (4 experts: A,C,D,W) & 12.691 & 8.707 \\
& FedProx Shared-expert PFL-MoE & 12.698 & 8.779 \\
& Private-expert PFL-MoE (top-$1$ routing) & 12.826 & 8.673 \\
& Private-expert PFL-MoE (4 experts: F,C,D,W) & 12.829 & 8.531 \\
& Shared-expert PFL-MoE & 12.837 & 8.906 \\
& Private-expert PFL-MoE (top-$5$ routing) & 12.994 & 8.749 \\
& Private-expert PFL-MoE (4 experts: F,A,C,D) & 12.997 & 8.586 \\
& Shared-expert PFL-MoE (random CS) & 13.100 & 8.832 \\
& PFL without MoE & 13.260 & \textbf{8.466} \\
& PFL-LoRA & 14.144 & 9.830 \\
& FedProx Global FL-MoE & 14.163 & 10.124 \\
& Global FL-MoE & 14.212 & 9.873 \\
& Local MoE & 14.298 & 9.327 \\
& Global FL-LoRA & 15.407 & 11.416 \\
& Local LoRA & 15.766 & 10.638 \\
& Zero-shot TSFM & 38.869 & 32.649 \\

\bottomrule
\end{tabular}
\end{table}

We analyze forecasting accuracy, expert-sharing strategies,
communication cost, and the routing behavior defined in
Section~\ref{sec:routing_fusion} across the three TSFM
backbones. Table~\ref{tab:complete_ablation} reports the complete
forecasting and ablation results, including main baselines,
expert-bank ablations, routing-sparsity variants, optimizer
variants, client-selection variants, and LoRA baselines.

\begin{figure}[t]
    \centering
    \includegraphics[width=\linewidth]{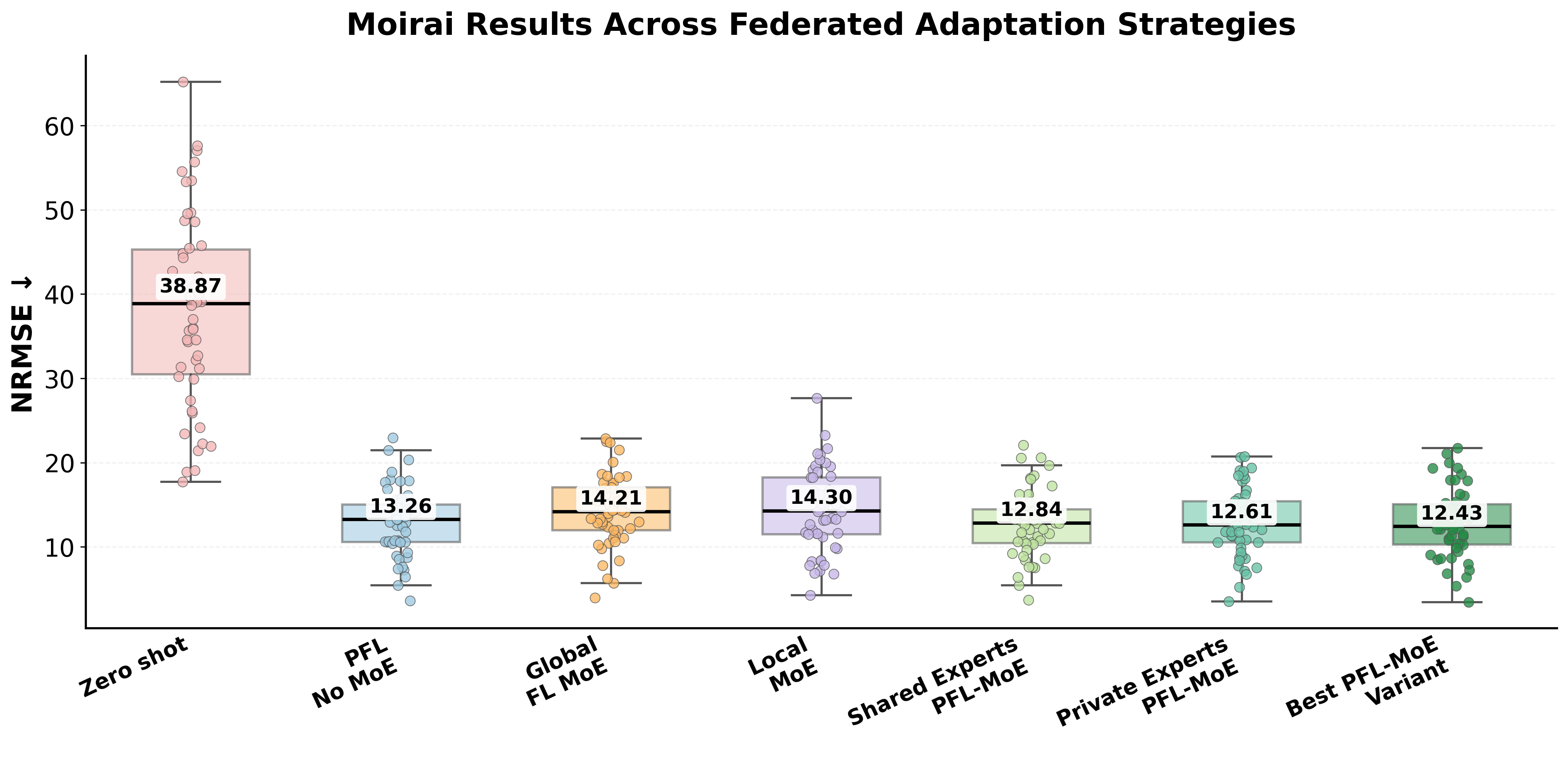}
    \caption{Per-client NRMSE distribution across 50 buildings for Moirai. 
    \textsc{PFL without MoE} denotes personalized federated adaptation without the MoE branch. 
    \textsc{Global FL-MoE} aggregates the trainable MoE adaptation globally, while 
    \textsc{Shared Experts PFL-MoE} and \textsc{Private Experts PFL-MoE} use shared and private expert banks, respectively. 
    Lower values indicate better forecasting performance.}
    \label{fig:moirai_nrmse_boxplot}
\end{figure}

To assess statistical significance, we performed paired Wilcoxon signed-rank
tests with Holm correction across building-level NRMSE and sMAPE values.
Private-expert PFL-MoE significantly outperformed Global FL-MoE across all
three backbones, whereas improvements over PFL without MoE were not
consistently significant after correction.

\begin{figure}[t]
    \centering
    \vspace{-6pt}
    \includegraphics[
        width=\columnwidth
    ]{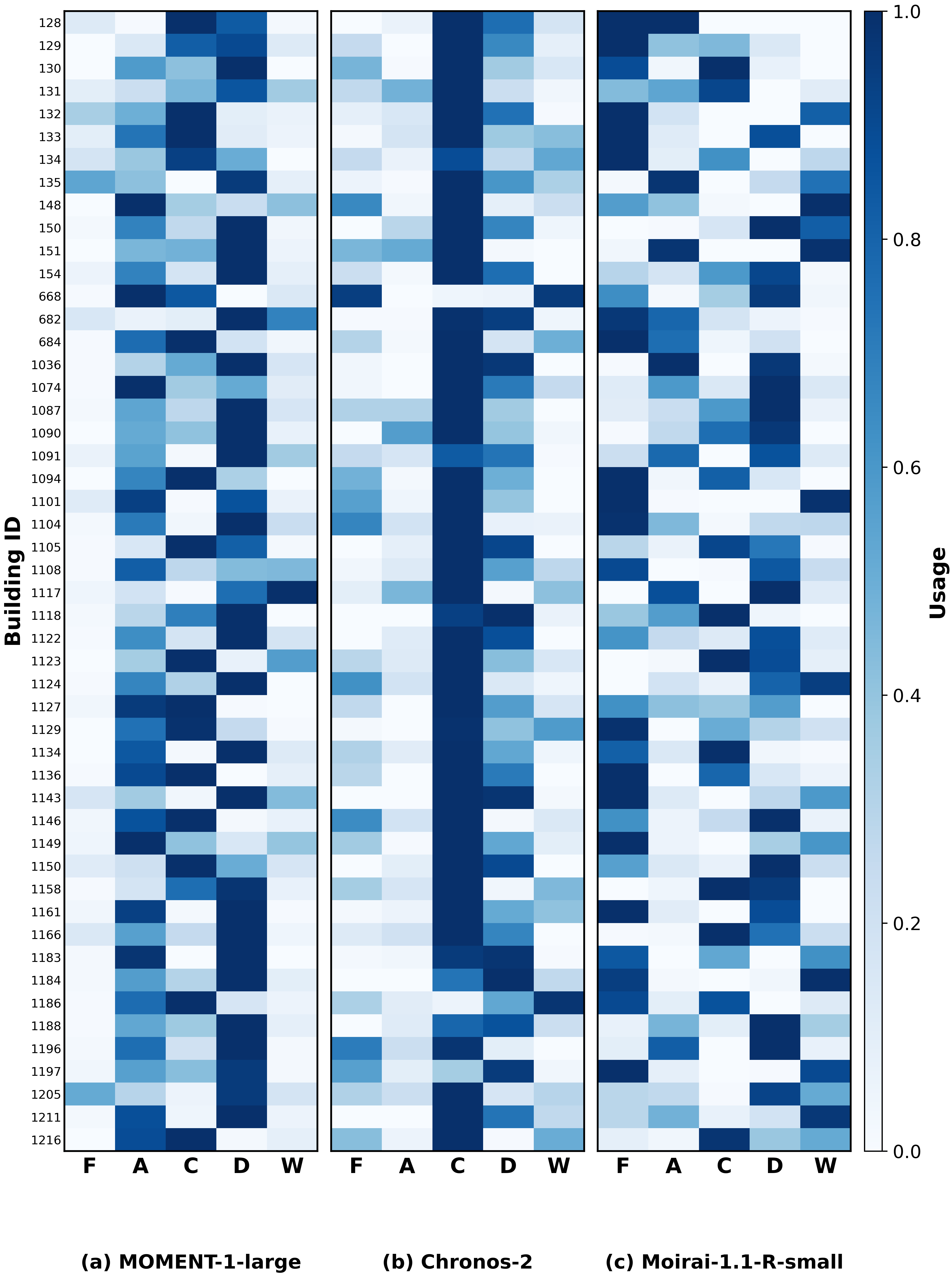}
    \vspace{-6pt}
    \caption{
    Expert-selection patterns for MOMENT-1-large, Chronos-2, and
    Moirai-1.1-R-small. Each row corresponds to one building client and each
    column corresponds to one expert. The color of a cell shows how often that
    expert was selected for that building's test windows by the top-$k$ router.
    MOMENT shows more building-specific expert choices, Chronos-2 selects a
    smaller set of experts more frequently, and Moirai uses the expert bank
    more evenly.
    }
    \label{fig:routing_fingerprint_hard_all_tsfms}
    \vspace{-10pt}
\end{figure}

\paragraph{Effect of personalization.}
Across all three backbones, the best personalized variant in
Table~\ref{tab:complete_ablation} reduces NRMSE relative to the corresponding
Global FL-MoE baseline by 8.2\% for MOMENT, 7.1\% for Chronos-2, and 12.5\%
for Moirai. These gains are computed within each backbone using Global
FL-MoE as the reference. Personalized variants generally outperform or remain competitive with Local
MoE. The statistically significant gains over Local MoE are observed for
Chronos-2 and Moirai, while MOMENT shows no significant improvement over Local
MoE after Holm correction.

\paragraph{Chronos-2.}
For Chronos-2, zero-shot performance is already very strong, indicating that
the pretrained representation is well aligned with the building-energy
forecasting task. A four-expert PFL-MoE variant achieves the best NRMSE,
while another four-expert variant obtains the best sMAPE. Personalized sparse adaptation generally outperforms Global FL-MoE and remains competitive with Local MoE.
However, the improvement over zero-shot forecasting remains modest,
suggesting that Chronos-2 already learns representations that transfer
effectively to building-energy forecasting. Consequently, adaptation has a
relatively limited impact on overall forecasting accuracy and primarily
affects expert selection and relative-error behavior. 

\paragraph{MOMENT.}
For MOMENT, top-$1$ routing achieves the best NRMSE, while private-expert
PFL-MoE obtains the best sMAPE. PFL without MoE and shared-expert PFL-MoE
remain competitive in NRMSE, suggesting that personalized adaptation already
captures much of the client-specific information required by this backbone.
The stronger sMAPE performance of private-expert variants indicates that
sparse expert routing mainly improves relative-error behavior rather than
substantially changing absolute forecasting accuracy.

\paragraph{Moirai.}
For Moirai, FedProx PFL-MoE with private experts achieves the best NRMSE,
with the four-expert FedAvg variant performing nearly identically. Both
outperform Global FL-MoE, Local MoE, PFL without MoE, and zero-shot
evaluation in terms of NRMSE. This suggests that Moirai benefits from
client-private expert transformations and that a single globally averaged
expert bank may be restrictive under heterogeneous building distributions.
However, PFL without MoE achieves the best sMAPE, indicating that the
optimal parameter-sharing strategy depends on the evaluation metric.

Figure~\ref{fig:moirai_nrmse_boxplot} further shows that private-expert
PFL-MoE shifts the per-client NRMSE distribution below both Global FL-MoE
and Local MoE for Moirai, indicating that the improvement is not driven by
only a few buildings. Similar per-client NRMSE distributions for MOMENT and Chronos-2 are shown in Appendix
Figure~\ref{fig:moment_chronos_nrmse_boxplots}. The larger personalization gains observed for Moirai
further suggest that client-specific temporal transformations are
particularly important for this backbone and may be difficult to represent
using a single globally shared expert bank.

\paragraph{Shared versus private experts.}
Expert sharing is not uniformly beneficial. Private experts improve sMAPE over
shared experts for MOMENT and Chronos-2 and give the strongest Moirai NRMSE
among the main FedAvg partitions. Shared experts remain competitive for
MOMENT NRMSE, but are weaker for Moirai, suggesting that globally averaged
expert banks can suppress client-specific temporal specialization.

\paragraph{Communication cost.}
A practical advantage of private experts is reduced communication by limiting which trainable parameters are
shared across clients. We compare communication across parameter-sharing
strategies rather than across the best-performing accuracy rows. Communication
is computed as $2RK|\theta_s|\times 4$ bytes, where $|\theta_s|$ is the number
of shared trainable parameters, $R=15$ rounds, and $K=10$ denotes the number of selected clients per round. This accounting includes only shared-parameter transmission in the
download and upload directions, and excludes optimizer states and client-private
parameters. Compared with shared-expert PFL-MoE, private-expert PFL-MoE reduces
communication by 66.3\%, 58.6\%, and 66.8\% for MOMENT, Chronos-2, and Moirai,
respectively, while adding only 8.2\%, 6.3\%, and 8.3\% overhead over PFL
without MoE. Table~\ref{tab:comm_cost_main} reports the corresponding
shared-parameter counts and total communication. While communication is reduced through private experts, this strategy increases client-side storage because each client maintains its own expert bank. A communication-accuracy trade-off visualization for all three
backbones is shown in Figure~\ref{fig:comm_tradeoff} in the appendix. 
\paragraph{Routing behavior.}
Figure~\ref{fig:routing_fingerprint_hard_all_tsfms} shows that the three
backbones induce distinct expert-usage regimes. MOMENT exhibits more
building-specific expert preferences, Chronos-2 concentrates selections on a
small subset of experts, and Moirai shows a more distributed usage pattern.
These differences suggest that pretrained TSFM backbones organize
building-level heterogeneity differently in the sparse-routing space.

\begin{table}[t]
\centering
\scriptsize
\setlength{\tabcolsep}{3pt}
\renewcommand{\arraystretch}{0.9}
\caption{
Communication cost under coverage-based client selection. Communication is
computed for bidirectional FP32 transmission of shared parameters, using 15
rounds and 10 selected clients per round.
}
\label{tab:comm_cost_main}
\begin{tabular}{llcc}
\toprule
\textbf{Backbone} & \textbf{Strategy} & \textbf{Shared Params (M)} & \textbf{Comm. (GB)} \\
\midrule
MOMENT & PFL without MoE & 12.85 & 15.42 \\
& Shared experts & 41.28 & 49.53 \\
& Private experts & 13.90 & 16.68 \\
\midrule
Chronos-2 & PFL without MoE & 9.44 & 11.33 \\
& Shared experts & 24.22 & 29.06 \\
& Private experts & 10.03 & 12.04 \\
\midrule
Moirai & PFL without MoE & 1.77 & 2.12 \\
& Shared experts & 5.77 & 6.93 \\
& Private experts & 1.92 & 2.30 \\
\bottomrule
\end{tabular}
\vspace{-6pt}
\end{table}

\paragraph{Limitations.}
This study focuses on 50 non-residential buildings from one building-energy
benchmark and on hourly univariate electricity forecasting. Broader validation
across climates, building types, multivariate forecasting settings, and larger
client populations is needed. 

\section{Conclusion}
\label{sec:conclusion}

We studied personalized federated adaptation of pretrained TSFMs for
short-term building energy forecasting. Across 50 non-IID building clients and
three TSFM backbones, personalized adaptation consistently improves over Global FL-MoE and is generally competitive with or better than Local MoE, with statistically significant gains over Local MoE for Chronos-2 and Moirai, demonstrating that federated TSFM adaptation benefits
from combining cross-client knowledge transfer with client-specific
specialization.

The strongest adaptation strategy depends on both the backbone and evaluation
metric. Chronos-2 exhibits strong zero-shot performance and is most sensitive
to expert-bank composition; MOMENT shows relatively small NRMSE differences
but benefits from sparse routing and private experts in terms of sMAPE; and
Moirai achieves the best NRMSE with private-expert FedProx PFL-MoE.
Expert-selection patterns further reveal that different TSFMs organize client
heterogeneity differently, while private experts reduce communication by
limiting the shared parameter set.

More broadly, our findings suggest that federated adaptation of foundation
models should be designed jointly with the characteristics of the underlying
pretrained backbone. The degree of personalization that is beneficial for
sparse expert adaptation varies substantially across TSFMs, indicating that
backbone-aware federated adaptation may be as important as client-aware
adaptation.

The backbone-dependent behavior observed across MOMENT, Chronos-2, and
Moirai suggests that there is no universally optimal personalization
strategy for federated TSFM adaptation. Instead, the degree of
personalization may itself need to be learned. Future work could develop
adaptive expert-sharing mechanisms that dynamically determine which
experts should be shared globally and which should remain client-specific,
potentially combined with meta-learned adapters that enable rapid
adaptation to new clients. More broadly, our results indicate that
backbone-aware personalization may be a key design principle for
federated foundation-model adaptation beyond building energy forecasting.

\bibliographystyle{named}
\bibliography{ijcai26}
\FloatBarrier

\clearpage
\appendix
\section{Additional Analysis}
\label{app:additional_analysis}
\subsection{Temporal Expert Implementations}
\label{app:expert_implementations}

This section provides implementation details for the expert modules used in the proposed sparse Mixture-of-Experts (MoE) adaptation branch. Unless otherwise noted, the same expert formulations are
used across MOMENT, Chronos-2, and Moirai, with only
minor backbone-specific hyperparameter differences. The adapter residual update is described in Section~\ref{sec:adapter}, the routing formulation and sparse top-$k$ expert fusion are described in Section~\ref{sec:routing_fusion}, and the auxiliary routing losses are described in Section~\ref{sec:objective_training}. Therefore, this section focuses only on the expert-specific transformations used inside the MoE branch.

Let the final hidden representation produced by the pretrained TSFM backbone be
\begin{equation}
\mathbf{Z} \in \mathbb{R}^{B \times P \times D},
\end{equation}
where $B$ is the batch size, $P$ is the number of temporal patches or hidden tokens, and $D$ is the hidden dimension. Each expert receives the same hidden representation $\mathbf{Z}$ and returns an output of the same shape:
\begin{equation}
E_m(\mathbf{Z}) \in \mathbb{R}^{B \times P \times D}.
\end{equation}
Thus, all experts operate in the latent patch-token space rather than directly on raw time-series values.

All experts follow the same general structure:
\begin{equation}
E_m(\mathbf{Z}) =
\mathrm{OutProj}_m\left(
\Phi_m(\mathrm{LN}_m(\mathbf{Z}))
\right),
\end{equation}
where $\mathrm{LN}_m(\cdot)$ is expert-specific layer normalization, $\Phi_m(\cdot)$ is the expert-specific temporal transformation, and $\mathrm{OutProj}_m$ maps the transformed representation back to dimension $D$. The output projection of each expert is initialized with a small standard deviation so that the MoE branch initially applies only a small residual perturbation to the frozen backbone representation.

The final expert bank used in all experiments consists of
five experts: Fourier, Attention, 
Convolution, Decomposition, and Wavelet. 

\subsubsection{Fourier Expert}
\label{app:fourier_expert}

The Fourier expert captures periodic and frequency-domain structure in the latent patch representation. The Fourier expert applies a real-valued FFT along the patch dimension of the normalized representation.
\begin{equation}
\widehat{X} = \mathrm{rFFT}(X, \mathrm{dim}=P),
\qquad
\widehat{X} \in \mathbb{C}^{B \times N_f \times D},
\end{equation}
where
\begin{equation}
N_f = \left\lfloor \frac{P}{2} \right\rfloor + 1
\end{equation}
is the number of frequency bins.

The implementation uses two learnable frequency filters:
\begin{equation}
G_{\mathrm{low}}, G_{\mathrm{band}} \in \mathbb{R}^{N_f \times D}.
\end{equation}
The low-frequency filter is initialized to ones:
\begin{equation}
G_{\mathrm{low}}(f,d) = 1.
\end{equation}
where f indexes the frequency bins and d indexes the hidden dimension.

The band-pass filter is initialized with a Gaussian-shaped profile:
\begin{equation}
G_{\mathrm{band}}(f,d)
=
\exp \left(
-\frac{1}{2}
\left(
\frac{f-c}{w}
\right)^2
\right),
\end{equation}
where
\begin{equation}
c = \max\left(1,\left\lfloor \frac{N_f}{4} \right\rfloor\right),
\qquad
w = \max\left(1,\left\lfloor \frac{N_f}{8} \right\rfloor\right).
\end{equation}

The filtered spectra are
\begin{equation}
\widehat{X}_{\mathrm{low}}
=
\widehat{X} \odot G_{\mathrm{low}},
\qquad
\widehat{X}_{\mathrm{band}}
=
\widehat{X} \odot G_{\mathrm{band}}.
\end{equation}
Both filtered representations are transformed back to the patch domain using the inverse real FFT:
\begin{equation}
X_{\mathrm{low}}
=
\mathrm{irFFT}(\widehat{X}_{\mathrm{low}}),
\qquad
X_{\mathrm{band}}
=
\mathrm{irFFT}(\widehat{X}_{\mathrm{band}}).
\end{equation}

The two components are concatenated along the hidden dimension and mixed:
\begin{equation}
U_F
=
[X_{\mathrm{low}}; X_{\mathrm{band}}] W_F + b_F,
\qquad
W_F \in \mathbb{R}^{2D \times D}.
\end{equation}
The final expert output is
\begin{equation}
E_F(Z)
=
\mathrm{OutProj}_F(\mathrm{Dropout}(U_F)).
\end{equation}

The Fourier expert is designed to capture periodic patterns such as daily or weekly load cycles. The low-frequency filter emphasizes smooth long-range variation, while the band-pass filter focuses on intermediate-frequency changes. Since both filters are learnable, the model can adapt the frequency emphasis during training.

\subsubsection{Attention Expert}
\label{app:attention_expert}

The attention expert models long-range dependencies between latent patch tokens. Given
\begin{equation}
X = \mathrm{LN}_{A}(Z),
\end{equation}
multi-head self-attention is applied over the patch dimension:
\begin{equation}
Q = XW_Q,
\qquad
K = XW_K,
\qquad
V = XW_V.
\end{equation}
For each attention head, the attention output is
\begin{equation}
\mathrm{Attn}(Q,K,V)
=
\mathrm{softmax}
\left(
\frac{QK^\top}{\sqrt{d_h}}
\right)V,
\end{equation}
where $d_h$ is the head dimension. The outputs from all heads are concatenated and projected back to dimension $D$:
\begin{equation}
E_A(Z)
=
OutProj_A
\Bigl(
Dropout(MHA(Q,K,V))
\Bigr).
\end{equation}

The attention expert captures non-local interactions between patches. In building energy forecasting, this can help relate distant parts of the input window, such as repeated occupancy patterns, morning ramp-up behavior, or evening shutdown behavior.

\subsubsection{Convolution Expert}
\label{app:conv_expert}

The convolution expert captures local and medium-range temporal variation in the latent patch sequence. Given
\begin{equation}
X = \mathrm{LN}_{C}(Z),
\end{equation}
the tensor is transposed so that the hidden dimension becomes the channel dimension:
\begin{equation}
X^\top \in \mathbb{R}^{B \times D \times P}.
\end{equation}
A depthwise dilated one-dimensional convolution is then applied:
\begin{equation}
C
=
\mathrm{DWConv1D}_{k=3,\delta=2}(X^\top),
\end{equation}
where $k=3$ is the kernel size and $\delta=2$ is the dilation factor. Depthwise convolution applies a separate temporal filter to each hidden channel. The output is then transposed back to shape $B \times P \times D$.

A pointwise feed-forward transformation is applied:
\begin{equation}
U_C
=
W_2 \,
\mathrm{Dropout}
\left(
\mathrm{GELU}(CW_1 + b_1)
\right)
+
b_2,
\end{equation}
where
\begin{equation}
W_1 \in \mathbb{R}^{D \times 2D},
\qquad
W_2 \in \mathbb{R}^{2D \times D}.
\end{equation}
The final expert output is
\begin{equation}
E_C(Z) = U_C.
\end{equation}

The convolution expert uses a dilated convolution to capture short- and medium-range temporal changes. Dilation increases the receptive field without substantially increasing the number of parameters, making it useful for local load fluctuations, ramp-up behavior, and ramp-down behavior.

\subsubsection{Decomposition Expert}
\label{app:decomposition_expert}

The decomposition expert separates the latent patch representation into a smooth trend component and a residual component. Given
\begin{equation}
X = \mathrm{LN}_{D}(Z),
\end{equation}
a frozen depthwise moving-average convolution is applied along the patch dimension:
\begin{equation}
X_{\mathrm{trend}} = \mathrm{AvgConv1D}(X).
\end{equation}
The moving-average kernel is initialized as
\begin{equation}
w_i = \frac{1}{K_{MA}}, \quad i=1,\ldots,K_{MA}.
\end{equation}
In Chronos-2, the moving-average kernel size is selected as

\[
K_{\text{MA}}=\max\left(5,\left\lfloor\frac{P}{4}\right\rfloor\right),
\]

and adjusted to be odd when required. For MOMENT and
Moirai, a fixed kernel size \(K_{\text{MA}}=9\) is used. The convolution
uses replicate padding and is kept frozen, so it acts as a
deterministic trend extractor.

The residual component is
\begin{equation}
X_{\mathrm{res}} = X - X_{\mathrm{trend}}.
\end{equation}
The trend and residual components are concatenated:
\begin{equation}
C_D =
[X_{\mathrm{trend}}; X_{\mathrm{res}}]
\in \mathbb{R}^{B \times P \times 2D}.
\end{equation}
A fusion projection maps the concatenated representation back to dimension $D$:
\begin{equation}
U_D = C_D W_{\mathrm{fuse}} + b_{\mathrm{fuse}},
\qquad
W_{\mathrm{fuse}} \in \mathbb{R}^{2D \times D}.
\end{equation}
The final expert output is produced using a pointwise feed-forward block:
\begin{equation}
E_D(Z)
=
W_2 \,
\mathrm{Dropout}
\left(
\mathrm{GELU}(U_D W_1 + b_1)
\right)
+
b_2.
\end{equation}

The decomposition expert introduces a trend--residual inductive bias. The trend component represents slowly varying latent behavior, while the residual component captures deviations from that trend. This is useful for building energy data, where consumption often contains both smooth base-load behavior and sharper occupancy- or equipment-driven changes.

Importantly, this decomposition is applied to the backbone's latent patch representation, not directly to the raw meter readings.

\subsubsection{Wavelet Expert}
\label{app:wavelet_expert}

The wavelet expert captures multi-resolution temporal structure using a Haar-style decomposition over the patch dimension. Given
\begin{equation}
X = \mathrm{LN}_{W}(Z),
\end{equation}
the expert repeatedly applies the Haar transform. At each level, the current sequence is split into even and odd patch positions:
\begin{equation}
X_{\mathrm{even}} = X_{0::2},
\qquad
X_{\mathrm{odd}} = X_{1::2}.
\end{equation}
The approximation and detail coefficients are computed as
\begin{equation}
A =
\frac{X_{\mathrm{even}} + X_{\mathrm{odd}}}{\sqrt{2}},
\qquad
\Delta =
\frac{X_{\mathrm{even}} - X_{\mathrm{odd}}}{\sqrt{2}}.
\end{equation}
The approximation $A$ is passed to the next level, while the detail component $\Delta$ is stored.

The number of decomposition levels is
\begin{equation}
L_W
=
\min
\left(
3,
\left\lfloor \log_2(\max(P,2)) \right\rfloor
\right).
\end{equation}
If the sequence length is odd at a level, the final unmatched patch is dropped for that level.

Each detail component is upsampled back to the original patch length $P$ using nearest-neighbor interpolation:
\begin{equation}
\widetilde{\Delta}^{(\ell)}
=
\mathrm{Upsample}(\Delta^{(\ell)}, P).
\end{equation}
The final approximation is also upsampled:
\begin{equation}
\widetilde{A}^{(L_W)}
=
\mathrm{Upsample}(A^{(L_W)}, P).
\end{equation}
All components are concatenated:
\begin{equation}
C_W =
[
\widetilde{\Delta}^{(1)};
\widetilde{\Delta}^{(2)};
\ldots;
\widetilde{\Delta}^{(L_W)};
\widetilde{A}^{(L_W)}
]
\in
\mathbb{R}^{B \times P \times (L_W+1)D}.
\end{equation}
A fusion projection maps this representation back to dimension $D$:
\begin{equation}
U_W = C_W W_{\mathrm{fuse}} + b_{\mathrm{fuse}},
\qquad
W_{\mathrm{fuse}} \in \mathbb{R}^{(L_W+1)D \times D}.
\end{equation}
The final expert output is
\begin{equation}
E_W(Z)
=
W_2 \,
\mathrm{Dropout}
\left(
\mathrm{GELU}(U_W W_1 + b_1)
\right)
+
b_2.
\end{equation}

The wavelet expert captures multi-resolution information. Fine-level detail coefficients represent short-term local changes, while deeper approximations represent coarser temporal structure. This is useful because building energy demand can contain both abrupt usage changes and slower long-term patterns.
\begin{table*}[t]
\centering
\caption{Parameter partitioning across adaptation strategies. Shared parameters are aggregated through federated learning, while private parameters remain local to each client.}
\label{tab:parameter_partitioning}
\small
\setlength{\tabcolsep}{10pt}

\begin{tabular}{p{0.22\textwidth}p{0.36\textwidth}p{0.36\textwidth}}
\toprule
\textbf{Method} & \textbf{Shared across clients} & \textbf{Private per client} \\
\midrule

Global FL-MoE &
Final backbone-side block, expert bank, router, residual gate, adapter output projection $\mathbf{W}_o$, and forecasting head &
None \\
\addlinespace

Local MoE &
None &
Final backbone-side block, expert bank, router, residual gate, adapter output projection $\mathbf{W}_o$, and forecasting head \\
\addlinespace

PFL without MoE &
Final backbone-side block &
Forecasting head \\
\addlinespace

Shared-expert PFL-MoE &
Final backbone-side block, expert bank, and adapter output projection $\mathbf{W}_o$ &
Router, residual gate, and forecasting head \\
\addlinespace

Private-expert PFL-MoE &
Final backbone-side block and adapter output projection $\mathbf{W}_o$ &
Router, residual gate, expert bank, and forecasting head \\
\bottomrule

\end{tabular}
\end{table*}

\begin{figure*}[t]
\centering
\includegraphics[width=0.32\linewidth]{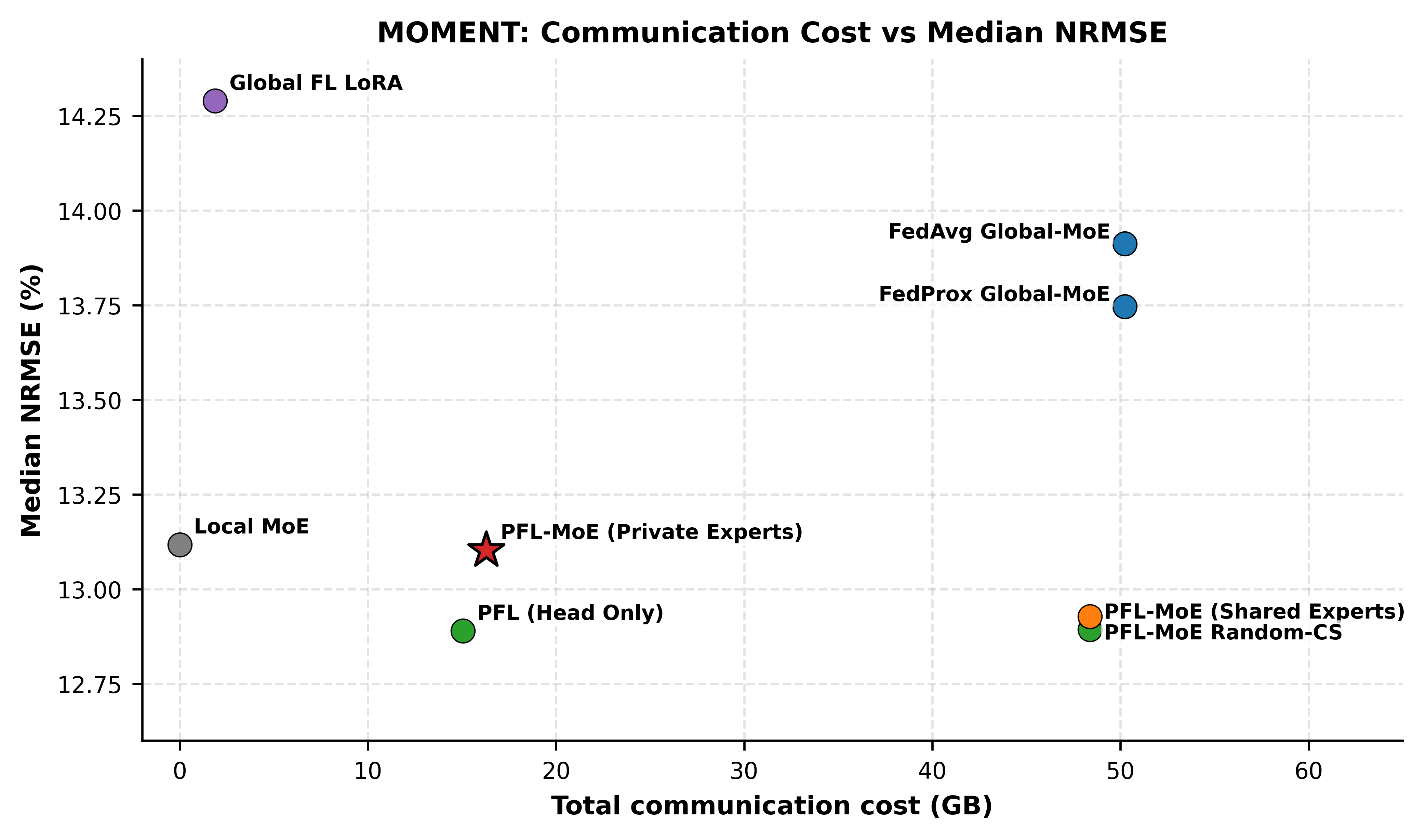}
\hfill
\includegraphics[width=0.32\linewidth]{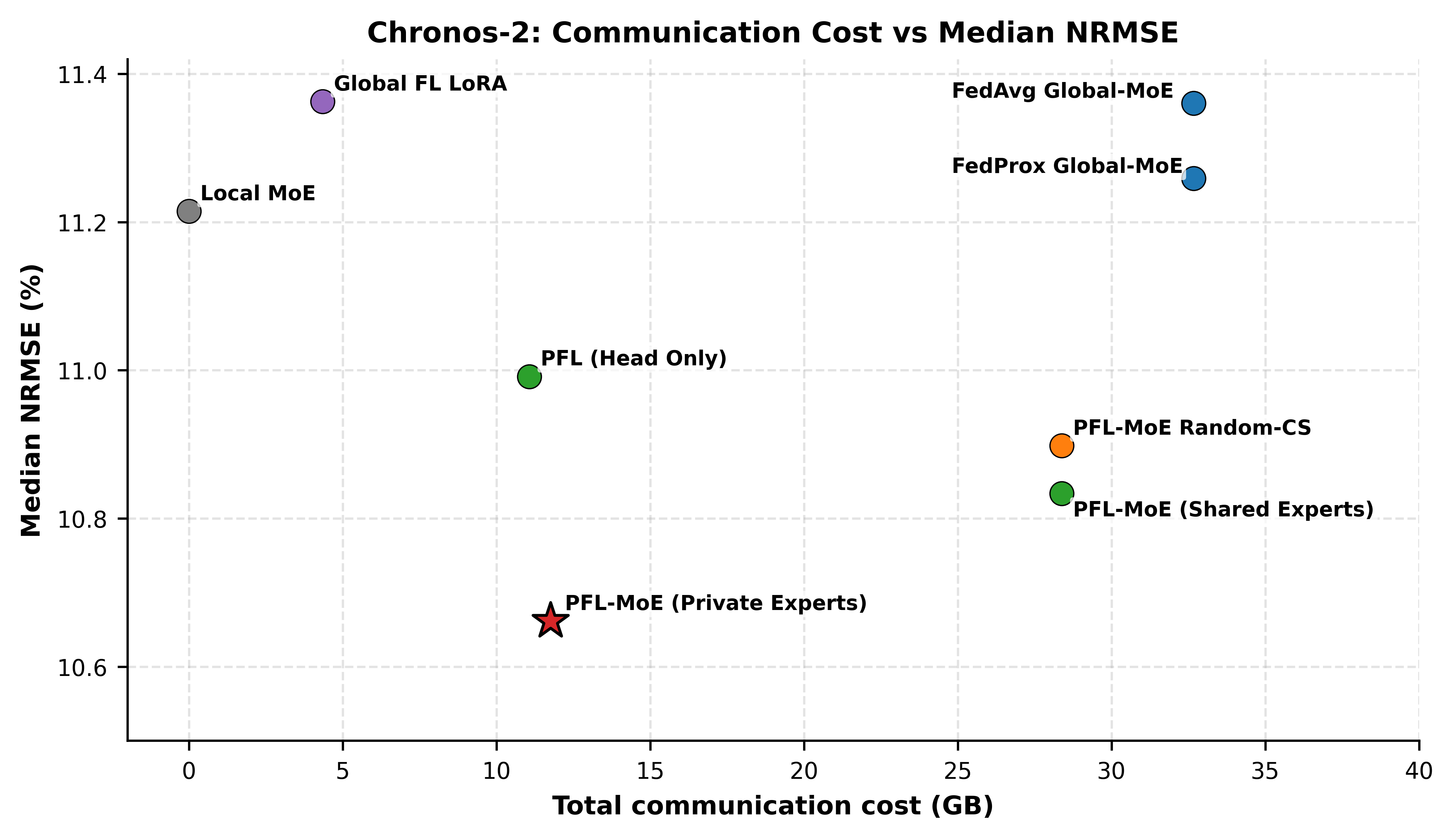}
\hfill
\includegraphics[width=0.32\linewidth]{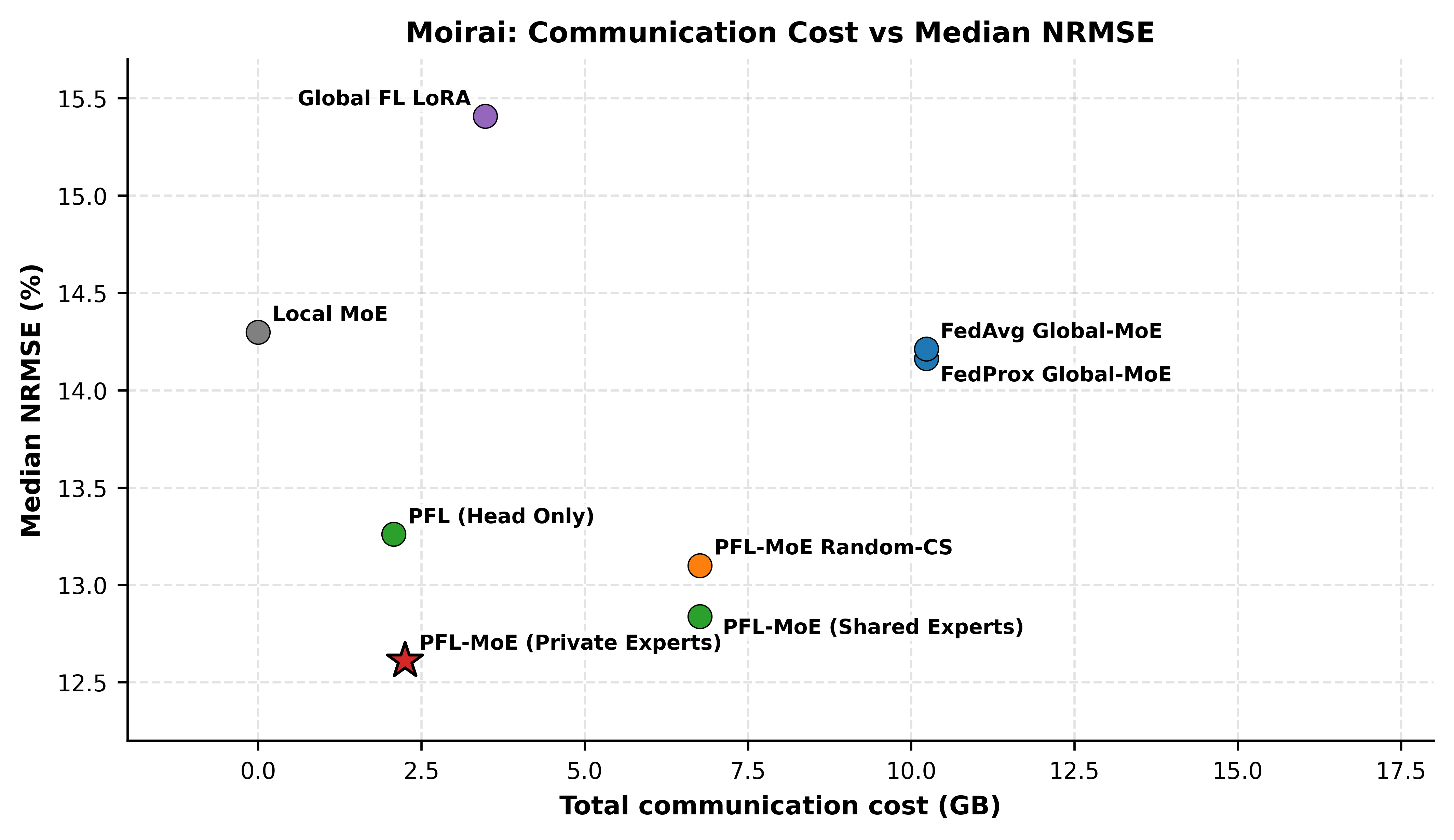}
\caption{
Communication cost versus median NRMSE across federated adaptation strategies for (a) MOMENT-1-large, (b) Chronos-2, and (c) Moirai-1.1-R-small. Lower-left regions correspond to preferable operating points with lower communication and lower forecasting error. Across all three backbones, personalized federated adaptation improves the communication--accuracy trade-off relative to globally shared FL-MoE. Private-expert PFL-MoE consistently achieves a favorable
communication--accuracy trade-off, attaining competitive or superior
forecasting performance while substantially reducing communication by
limiting expert parameters to client-private storage.
}
\label{fig:comm_tradeoff}
\end{figure*}

\FloatBarrier
\section{Training and Federated Learning Hyperparameters}
\label{app:hyperparameters}

\begin{table}[h]
\centering
\caption{
Training and federated learning hyperparameters shared across experiments.
Backbone-specific optimization settings follow the default configurations used
for each TSFM backbone.
}  
\label{tab:hyperparameters}
\scriptsize
\setlength{\tabcolsep}{3pt}
\renewcommand{\arraystretch}{0.95}
\begin{tabularx}{\columnwidth}{@{}lY@{}}
\toprule
\textbf{Parameter} & \textbf{Value} \\
\midrule
Context length ($L$) & 168 \\
Forecast horizon ($H$) & 24 \\
Communication rounds & 15 \\
Client participation rate & 20\% \\
Local epochs & 5 \\
Top-$k$ routing & 2 \\
Number of experts & 5 \\
Optimizer & AdamW \\
Batch size & 8 \\
Weight decay & $1\times10^{-4}$ \\
$\lambda_{\mathrm{lb}}$ &
$1\times10^{-4}$ for MOMENT and Chronos-2; $1\times10^{-3}$ for Moirai \\
$\lambda_z$ &
$1\times10^{-5}$ for MOMENT and Chronos-2; $1\times10^{-4}$ for Moirai \\
Initial gate logit ($\ell_\alpha$)
& 0 (MOMENT, Chronos-2), -1 (Moirai) \\
FedProx coefficient $\mu$ & 0.01\\
Training window stride & 1 \\
\bottomrule
\end{tabularx}
\vspace{-4pt}
\end{table}
\FloatBarrier

\subsection{Statistical Significance Testing}
\label{app:wilcoxon_tests}

We additionally evaluate whether the observed improvements of the proposed private-expert PFL-MoE method are statistically significant across building clients. Since forecasting performance is computed for the same set of buildings under each adaptation strategy, we use a paired non-parametric Wilcoxon signed-rank test. Each building contributes one paired score for the compared methods. The test is applied separately to the two primary evaluation metrics, NRMSE and sMAPE.

For each TSFM backbone, we compare private-expert PFL-MoE against three baselines: Global FL-MoE, Local MoE, and PFL without MoE. Since lower NRMSE and sMAPE values indicate better forecasting performance, the one-sided alternative hypothesis is
\begin{equation}
H_1: e_{\mathrm{baseline}} > e_{\mathrm{private\text{-}PFL\text{-}MoE}},
\end{equation}
where $e$ denotes the building-level metric value. Thus, a positive median improvement indicates that private-expert PFL-MoE reduces the corresponding error metric relative to the baseline.

For each backbone, we conduct six paired tests, corresponding to three baselines and two metrics. We therefore report both the raw Wilcoxon $p$-value and the Holm-Bonferroni adjusted $p$-value. The Holm--Bonferroni procedure is applied independently for each backbone to
control the family-wise error rate across the six hypothesis tests
(three baselines $\times$ two metrics). A comparison is considered
statistically significant when the Holm-adjusted $p$-value is below 0.05.
\begin{table*}[t]
\centering
\scriptsize
\caption{Paired Wilcoxon signed-rank tests comparing \emph{Private-expert PFL-MoE} with the main baselines across 50 building clients. Tests are performed on building-level NRMSE and sMAPE values. Holm-adjusted $p$-values correct for the six tests performed for each backbone.}
\label{tab:wilcoxon_appendix}

\resizebox{\textwidth}{!}{
\begin{tabular}{llcccc}
\toprule
\textbf{Backbone} &
\textbf{Comparison} &
\textbf{Metric} &
\textbf{Improved Clients} &
\textbf{Wilcoxon $p$} &
\textbf{Holm $p$} \\
\midrule

& Private-expert PFL-MoE vs Global FL-MoE & NRMSE & 33/50 & 0.0034 & 0.0168 \\
& Private-expert PFL-MoE vs Global FL-MoE & sMAPE & 31/50 & 0.0012 & 0.0071 \\
\textbf{MOMENT}
& Private-expert PFL-MoE vs Local MoE & NRMSE & 17/50 & 0.9297 & 1.0000 \\
& Private-expert PFL-MoE vs Local MoE & sMAPE & 24/50 & 0.2992 & 0.8977 \\
\midrule

& Private-expert PFL-MoE vs Global FL-MoE & NRMSE & 37/50 & $6.84 \times 10^{-6}$ & \textbf{$3.42 \times 10^{-5}$} \\
& Private-expert PFL-MoE vs Global FL-MoE & sMAPE & 42/50 & $9.34 \times 10^{-8}$ & \textbf{$5.60 \times 10^{-7}$} \\
\textbf{Chronos-2}
& Private-expert PFL-MoE vs Local MoE & NRMSE & 34/50 & $4.40 \times 10^{-4}$ & 0.0013 \\
& Private-expert PFL-MoE vs Local MoE & sMAPE & 36/50 & $1.49 \times 10^{-5}$ & $5.95 \times 10^{-5}$ \\
\midrule

& Private-expert PFL-MoE vs Global FL-MoE & NRMSE & 40/50 & $1.52 \times 10^{-7}$ & \textbf{$7.60 \times 10^{-7}$} \\
& Private-expert PFL-MoE vs Global FL-MoE & sMAPE & 43/50 & $4.89 \times 10^{-9}$ & \textbf{$2.93 \times 10^{-8}$} \\
\textbf{Moirai}
& Private-expert PFL-MoE vs Local MoE & NRMSE & 37/50 & $1.34 \times 10^{-5}$ & \textbf{$5.38 \times 10^{-5}$} \\
& Private-expert PFL-MoE vs Local MoE & sMAPE & 34/50 & $1.26 \times 10^{-4}$ & $3.77 \times 10^{-4}$ \\
\bottomrule
\end{tabular}
}
\end{table*}

The results show that private-expert PFL-MoE significantly improves over Global FL-MoE for all three TSFM backbones on both NRMSE and sMAPE after Holm correction. This supports the importance of personalized sparse adaptation compared with fully global MoE aggregation. Private-expert PFL-MoE also significantly improves over Local MoE for Chronos-2 and Moirai on both metrics, indicating that federated transfer remains beneficial compared with isolated local training for these backbones. For MOMENT, the improvement over Local MoE is not statistically significant.

Compared with PFL without MoE, private-expert PFL-MoE does not achieve statistically significant improvements after Holm correction for any of the three backbones. This suggests that the main statistically robust gain comes from personalized federated adaptation relative to global or local MoE training, while the additional benefit of the sparse MoE branch over non-MoE personalization is backbone- and metric-dependent.

\FloatBarrier

\begin{figure*}[!t]
    \centering

    \begin{minipage}{0.48\textwidth}
        \centering
        \includegraphics[width=\linewidth]{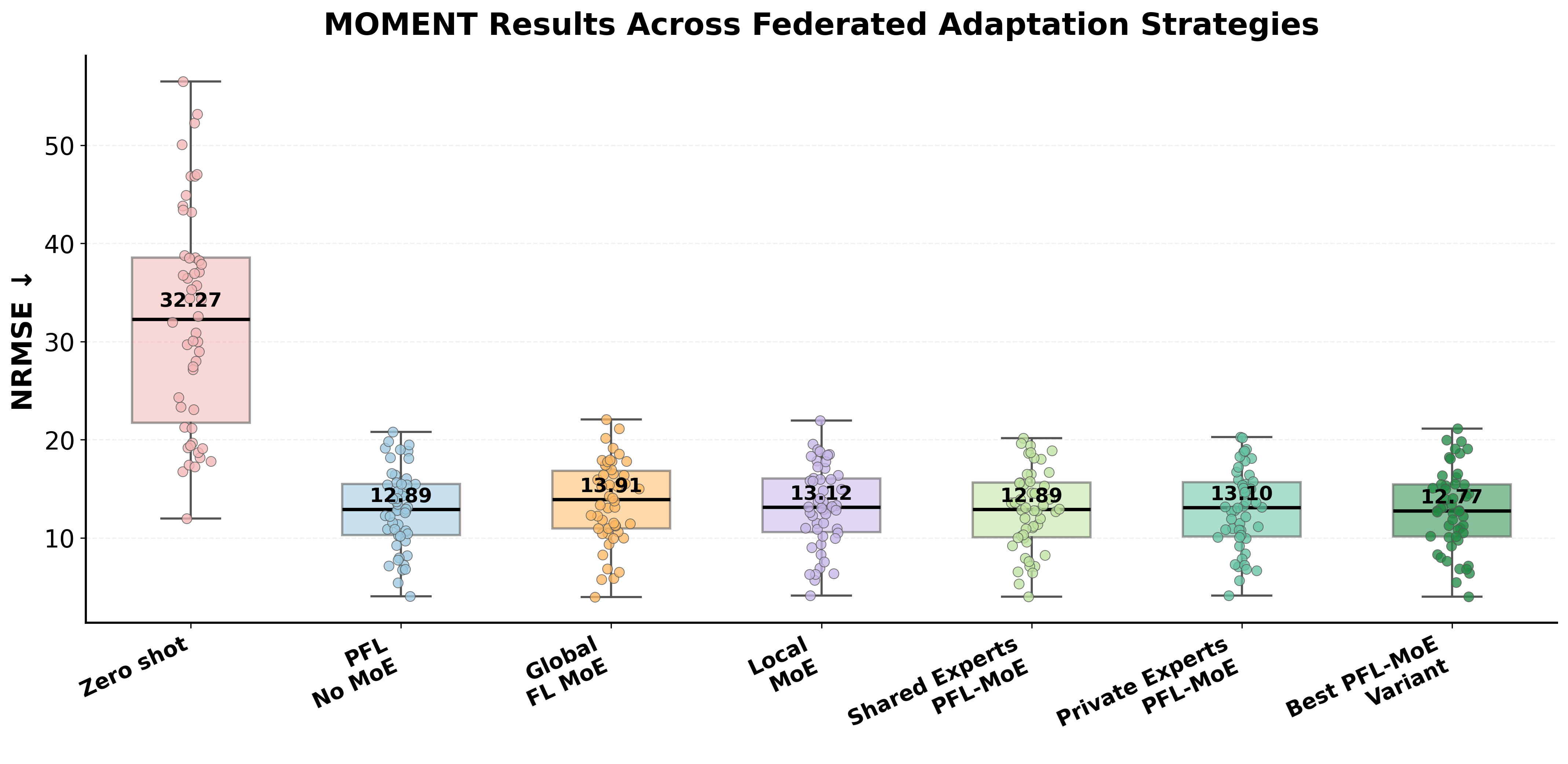}
        \vspace{-4pt}
        \textbf{(a) MOMENT-1-large}
    \end{minipage}
    \hfill
    \begin{minipage}{0.48\textwidth}
        \centering
        \includegraphics[width=\linewidth]{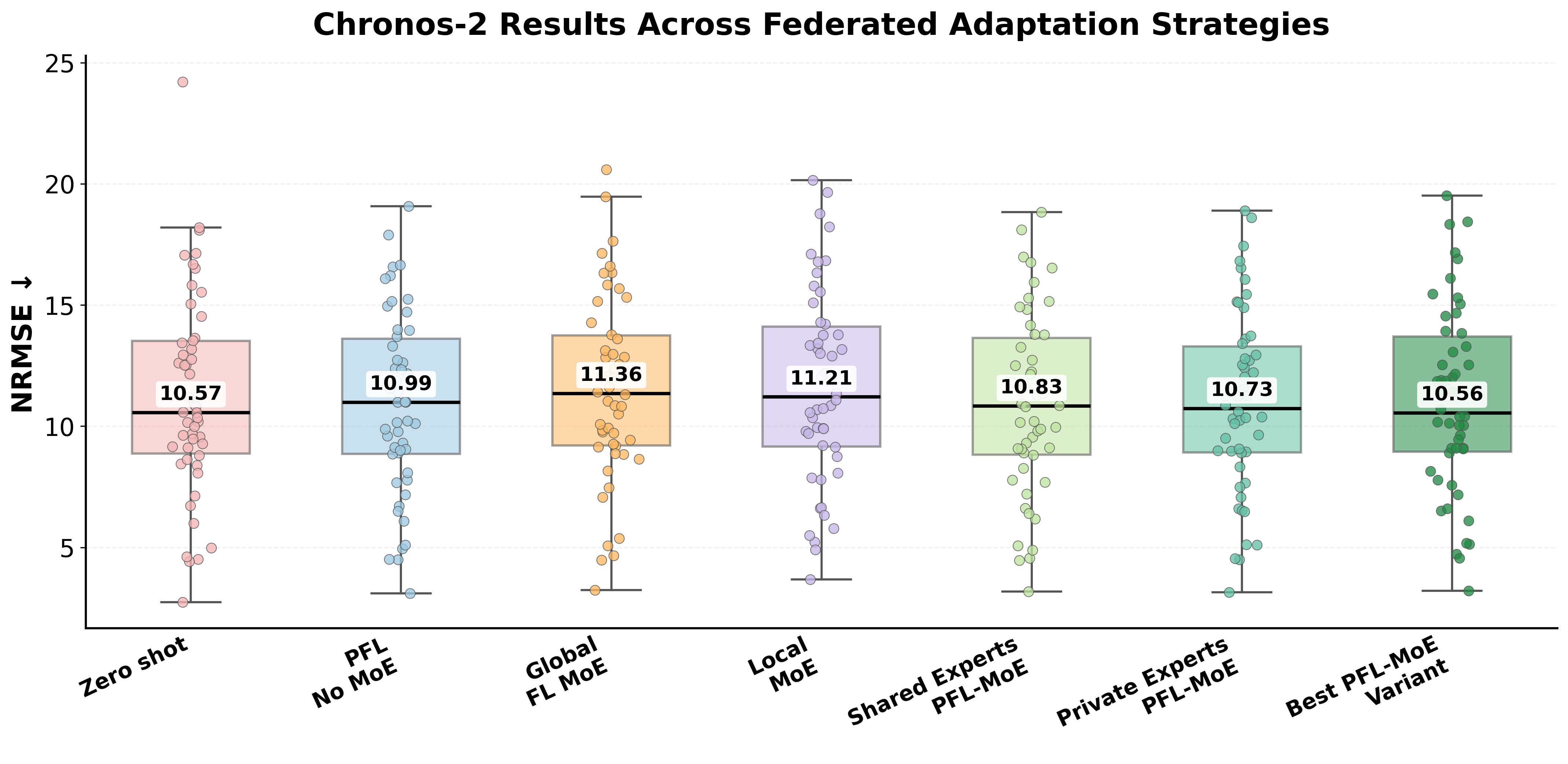}
        \vspace{-4pt}
        \textbf{(b) Chronos-2}
    \end{minipage}

    \caption{
    Per-client NRMSE distributions across 50 buildings for MOMENT-1-large
    and Chronos-2 under different adaptation strategies. The boxplots compare
    zero-shot inference, personalized FL without MoE, global FL-MoE, local MoE,
    shared-expert PFL-MoE, private-expert PFL-MoE, and the best-performing
    PFL-MoE ablation for each backbone. Lower NRMSE indicates better forecasting
    performance.
    }
    \label{fig:moment_chronos_nrmse_boxplots}
\end{figure*}

\FloatBarrier

\end{document}